# Nonverbal cues in human-robot interaction: A communication studies perspective

**JACQUELINE URAKAMI**

*Industrial Engineering & Economics*
*Tokyo Institute of Technology*

**KATIE SEABORN**

*Industrial Engineering & Economics*
*Tokyo Institute of Technology*

**ABSTRACT:** *Communication between people is characterized by a broad range of nonverbal cues. Transferring these cues into the design of robots and other artificial agents that interact with people may foster more natural, inviting, and accessible experiences. In this position paper, we offer a series of definitive nonverbal codes for human-robot interaction (HRI) that address the five human sensory systems (visual, auditory, haptic, olfactory, gustatory) drawn from the field of communication studies. We discuss how these codes can be translated into design patterns for HRI using a curated sample of the communication studies and HRI literatures. As nonverbal codes are an essential mode in human communication, we argue that integrating robotic nonverbal codes in HRI will afford robots a feeling of "aliveness" or "social agency" that would otherwise be missing. We end with suggestions for research directions to stimulate work on nonverbal communication within the field of HRI and improve communication between human and robots.*

**KEYWORDS:** *Communication studies, human-robot interaction, nonverbal communication, robotics, nonverbal codes*

## 1 INTRODUCTION

Communication studies has a long and rich history, with many of today's ideas being traceable back to scholars of ancient Greece and Rome. Communication is a complex process that can be defined in many ways. Before turning to applying what we know about human communication to human-robot interaction (HRI), we have to start with a useful definition of communication. In its broadest sense, communication is the discriminatory response of an organism to a stimulus [1]. According to Hauser [2], communication systems consists of three features: the signal source, a transmission channel, and a pool of perceivers. Thus, communication involves two or more communication partners who are either sending or receiving messages. In HRI, one of the communication partners is a machine; while often a social robot, the machine may be any kind of human-interfacing agent, interface, or artificial intelligence (AI)-infused environment, such as a voice assistant (VA), smart vehicle, or conversational user interface (CUI), to name a few[1]. Therefore, any definition of communication in HRI needs to account for non-human (and potentially non-humanlike) message transmission and communication behavior that is intentional or unintentional. Furthermore, definitions of communication differ in being sender- or receiver-oriented. Since self-agency and conscious intent in sending a message are questionable[2] in the case of a robot as the sender, a receiver-based perspective is more useful in HRI. Therefore, we define communication as *an interactive process whereby a receiver (i.e., a person) assigns meaning to one or more stimuli that are transmitted intentionally or unintentionally by a sender (i.e., a robot)*. With this definition in hand, we must now articulate what factors of human-human communication may be applied to human-robot communication and to what extent.

One key factor is nonverbal communication, which is generally defined as *any transfer of messages that does not involve the use of words* as *nonverbal communication [3].* This also includes sounds that are not words, such as backchanneling (“uh”, “ah”) as well as the pitch, tone, and intonation of the voice. Between humans, nonverbal communication represents a significant ratio of all face-to-face communication. Human conversation, for instance, is accompanied by a variety of gestures, postures, facial expressions, and eye movements, among other *nonverbal codes*. Nonverbal codes describe the variety of nonverbal stimuli, or *cues*, that occur within a communication context. Nonverbal cues can then be described as *any information within a nonverbal communication process to which a receiver assigns meaning, regardless of whether this meaning was transmitted intentionally or not by the sender.* This "unspoken language" is an integral part of the message and its transmission in human-human communication. As such, nonverbal cues may be useful in cases where robots interact directly with people, as well. Indeed, recent research in HRI (e.g., [4-8]) has acknowledged the importance of nonverbal cues, especially for novice users in social settings (e.g., [6]). Moreover, designers of robots often attempt to consider how the appearance of a robot affects human behavior, such as by creating friendly-looking, approachable social robots. Examples include human-sized social robots like Pepper (SoftBank/Aldebaran Robotics), small, toy-like robots with facial expressions for play and entertainment like Cozmo (Digital Dream Labs), and sturdy, intimidating surveillance robots like the K5 (Knightscope). Considering the growing interest in designing robots to use nonverbal cues when communicating with people, a timely and useful addition to help direct future research and practice would be an overview of common nonverbal codes established by the field of human communication. In particular, we need guidance on how these codes may be, and, where possible, have been, applied in HRI contexts as cues.

---

[1] For the sake of simplicity and with relevance to this venue, we will use the term "robot" from here onward. However, we do not suggest that this material is limited to conventional robot categories. Indeed, the term "robot" can be defined in various ways and applied broadly to technologies ranging from algorithms to autonomous agents, and we use it in this spirit.

[2] We fully expect that this view will become outdated with advances in AI, especially around machine agency and self-awareness. For now, however, it is an accurate take on reality.

To this end, we present an overview of nonverbal communication for HRI, grounded in human communications research and existing work on robots and other artificial agents. The goals of this position paper are threefold: 1) to present a definition of nonverbal communication in HRI based on human communication models; 2) to provide a framework of nonverbal codes derived from human communication studies; and 3) to demonstrate how these nonverbal codes may be and have been applied in HRI. We introduce a theoretical framework of robotic nonverbal cues based on their perceptibility through the human sensory systems outlining nonverbal cues that address vision, auditory perception, haptics, olfactory and gustatory perception. We begin with an explanation of our rationale for including nonverbal communication in HRI. We then describe the major characteristics of nonverbal communication from communication studies. Following this, we outline and describe the framework of common nonverbal codes that play an important role in human communication and are relevant for HRI. We discuss and critique examples of where these codes have already been studied, as well as provide ideas for future research and development. We end with an outlook on the next steps for explorations of nonverbal codes in HRI.

## 2 ADAPTING HUMAN NONVERBAL "CODES" FOR HRI

Many in robotics believe that robots can coexist as social agents alongside humans, but not necessarily as their social equals. How robots and other AI-based machines become socially embodied agents, and whether this state is static or circumstantial and unstable, remains an open question [9]. Some Japanese roboticists have argued that to become a social agent, robots should engage the world with sensitivity, sensibility, feeling, aesthetics, emotion, affection, and intuition [10] a *kokoro function*. Kokoro in Japanese has a complex meaning that can be translated as any of heart, spirit, and/or mind, integrating emotion, thinking, and intention, and describing the *inner world* or *essence* of a person (or in this case a robot). Thus, robot creators in Japan often refer to robots as belonging to a "third existence", where a robot is something between a living and non-living creature [11]. A question remains about how this *inner world* of a robot can be made accessible to human communication partners. Furthermore, people's reactions and internal states are often expressed through physiological changes that are visible to others. Interpretation of physiological cues are highly contextual. Facial flushing can indicate embarrassment, stress, happiness, excitement, anger, and so on. Heavy breathing can signal anxiety, even a panic attack, or be a natural state of exertion after intense exercise. These cues can add to the perceived "aliveness" of a robot and contribute to impressions of its kokoro function. Incorporating physiological cues within HRI contexts can enhance the perception of a robot as being “alive” or human-like. For example, Lütkebohle et al. [12] simulated blushing in the robot Flobi with an array of red LEDs in its cheeks, whereas Yoshida and Yonezawa [13] developed a stuffed bear-style robot that can breathe, has a heartbeat, has varying temperature, and other involuntary bodily movements related to physiological and psychological states.

In this position paper, we develop the idea of giving robots a "liveness" or "social agency" through the use of nonverbal communication modalities. In human communication, nonverbal codes are an essential mode of communication that, among other things, reveal a person's role and status (e.g., through clothing style), provide information about feelings and emotions (e.g., facial expressions, gestures), or offer redundancy in speech (e.g., gestures and body movements). Enriching robot communication with nonverbal codes derived from human communication could help human collaborators perceive robots as social agents improving mutual understanding and coordination. Explorations in nonverbal communication with robots has a history going back several decades [6, 14]. Roboticists have sought to create intuitive interactions between robots and people and increase human acceptability of robots [6, 15]. Nehaniv et al. [16] developed a taxonomy of gestures related to robotic arm movements. De Santis et al.[17] surveyed the HRI literature on physical HRI involving mechanical and control components. Previous research has looked at affective expressions in robotic body movements that did not include face or verbal components [18], or studied "tactile HRI," or the ability of robots to detect a person's physical touch [19]. In a social twist, Van Erp and Toet [20] produced ten guidelines for affective and interpersonal components of robotic touch. Kruse et al. [21] and later Rios-Martinez, Spalanzani, and Laugier [22] each covered social distance and other cues related to robot navigation of social spaces in their surveys. The literature review conducted by McColl et al. [14] was limited to human perception of robot intention through nonverbal gestures in decision-making scenarios. Admoni and Scassellati [23] focused on eye gaze in social HRI contexts. Saunderson and Nejat [6] focused on robotic motion and movement: kinesics, proxemics, haptics, and chronemics. Explorations along these lines are expected to continue as we continue to deploy robots in social contexts with people.

Yet, the vast majority of the work so far has focused on a limited range of nonverbal cues, often exploring a single cue in isolation rather than evaluating a range of cues, which is more representative of human models [6]. Additionally, there has been a strong technical focus, with far less work exploring how robots with nonverbal cues influence people's attitudes and behaviour [6]. In this paper, we take one step towards addressing these gaps and limitations by applying a comprehensive human communications studies perspective to the HRI context. We now turn to describing the nature of nonverbal communication among people, including the full range of cues and how they work together.

## 3 NONVERBAL COMMUNICATION CHARACTERISTICS

*Communication*, as we defined above, is an interactive process involving a signal source, a transmission channel, and an agent (or perceiver) acting on a stimulus. In HRI, either the agent or the source of the stimuli is a robot. What form this stimulus takes can vary. In the case of *nonverbal* communication, the stimuli do not take the form of speech. As such, we define *nonverbal communication in HRI* as any type of communication between people and robots that does not involve words, but includes nonverbal utterances. We build on a psychological model of communication that distinguishes between a sender who transmits a message and a receiver who decodes the message [24]. Communication is an interactive process, where messages are continuously exchanged between sender and receiver. Here, we take a simple approach, focusing on scenarios where the robot is the sender of nonverbal cues and a person is the receiver, having to decode and assign meaning to the cues transmitted by the robot. We define nonverbal cues as any stimuli a receiver (person) assigns meaning too, independently whether these nonverbal cues were transmitted intentionally or unintentionally by the sender (robot). Nonverbal communication differs from verbal communication in several important ways. Understanding the features of nonverbal communication that distinguish it from verbal communication is essential for the successful implementation of nonverbal cues in HRI, but it is also one of the greatest challenges. Transferring nonverbal gestures from humans to robots is likely to fail if the prerequisites for nonverbal codes are violated. In general, humans are accustomed to interpreting nonverbal codes in other humans and as such can be sensitive to robots that deviate from learned expectations. Nonverbal communication has several characteristics that distinguish it from other communication systems. We outline seven that we deem relevant for implementing nonverbal codes into robots.

### 3.1 Multiple Interpretations

*Nonverbal cues can be interpreted in different ways.* For example, a person moving away from a robot can be a sign of disinterest, fear, or surprise, or even signal an end to the interaction. In a similar way, nonverbal cues used by a robot can be interpreted by a person in different ways as well. Changing the eye color of a robot as done in a study by Koike et. al. [25] evoked a variety of different reactions in people. This poses a problem to the designer: how to make sure that the intended message has been received and understood, on both sides. Including redundant signals and using a variety of non-verbal codes can help to clarify the message.

### 3.2 Unintentional Behavior

*Nonverbal cues interpreted by a receiver may or may not have been part of the intended message*. The robot’s body shape and its movement might be constrained by what is technically feasible. Nonetheless, these features might be interpreted by a receiver as intentional nonverbal cues.

Also, it is difficult to convey different emotional expressions through the voice of a robot, such as sounding happy, sad, or concerned. An emotionless voice independent of what is said can be confusing for a receiver, and is often interpreted as a lack of empathy [26]. Additionally, when the affective quality of the voice does not match expectations, this can cause confusion or otherwise provide a negative experience [5].

### 3.3 Multiple Cues

*There are multiple, separate nonverbal cues that play a role in communication.* Clothing, facial expressions, and hand gestures all send out a message that may be interpreted individually or in concert. When the same message is sent via each cue, the combined impact of all cues is intensified. However, if cues are contradictory, then they can cause confusion and thereby undermine the impact of the message. For example, if a sender shakes their head left and right, a signal of "no" in most cultures, while also saying "yes" out loud. Exceptions exist, such as when a sender uses a certain code or sensory modality to send a message to a certain receiver and not others, or when a sender presents two cues simultaneously that are intended for different receivers. For example, a sender may be absorbed in playing a video game, with eyes glued to the screen, their whole body facing the game, and actively pounding away a game controller, while also carrying on a conversation about other matters with a receiver who is present but observing. On the technical side, nonverbal response and behaviour frameworks will need to be developed for robots or extended from existing, similar offerings, such as Lee and Marsella's [27] nonverbal behaviour generator for embodied conversational agents (ECA). Programming and markup languages developed for ECAs, such as the Behavior Markup Language (BML) [28], could be used to develop nonverbal expressions for robots in HRI contexts, as well.

### 3.4 Immediate Message Transmission

*Nonverbal cues are always being sent, intentionally or unintentionally, when we are face-to-face with a communication partner*. Basically, we cannot *not* communicate. As such, the mere presence of a robot involves transmission of messages that will likely be interpreted by a receiver. Furthermore, the presence or absence of specific behavior in the robot will likely be interpreted by a receiver. For example, mechanical noise generated by motors that move a robot's arms, fingers or legs become part of the nonverbal message and can create a disturbing effect.

### 3.5 Continuous

Verbal messages have a clear beginning and end, indicated by syllable, word, and sentence boundaries. *Nonverbal cues, on the other hand, are without discrete starting and ending points, but rather flow into one another.* It is difficult to isolate body movements and gestures, such as greeting someone or waving a hand. Similarly, facial expressions are not static; rather, they are a series of expressions that shift over time to indicate certain states. Indeed, creating continuous nonverbal cues is probably one of the biggest challenges in robotics. We already have some evidence of this, in that the isolated or separate nonverbal gestures that are often explored in robotics research or deployed in commercial robots tend to be difficult to interpret, leading to feelings of the robot being unnatural and uncanny [29].

### 3.6 Concrete

*Nonverbal cues are often concrete and representational rather than abstract and arbitrary*, compared to verbal messages. For example, a motion gesture can represent actual movements of objects or people in the real world, indicating a connection between form and meaning in nonverbal cues [30]. Therefore, nonverbal cues used in HRI need to be synchronized and especially match any speech content. Arbitrary and abstract gestures that are often preferred by experts when interacting with technical systems [31] should be avoided (as far as the technical limitations of robot motion allow) because they are difficult to interpret for novice users.

### 3.7 Contextual and Cultural

Even though some nonverbal cues, such as facial expressions for basic emotions in typical contexts (e.g., happiness, anger, fear, sadness, disgust, surprise) are interpreted similarly in most places in the world in modern times [32], *many nonverbal cues are only shared among specific cultural groups and within generational cohorts*. For example, [33] found that Japanese and Australians' evaluation of an android differed in perceived anthropomorphism, animacy, and perceived intelligence. Furthermore, simple cultural cues indicating in-group membership can help reduce unfavorable biases towards robots [34]. Therefore, nonverbal cues for robots need to be designed and/or carefully selected for the intended culture and cohort, as well.

## 4 THE NONVERBAL CODES

A variety of nonverbal codes have been discussed in communication studies. However, several factors need to be considered when selecting nonverbal codes for HRI. As mentioned above, we focus in this paper on a simple communication model whereby the robot is the sender and a human is the receiver. On the one hand, nonverbal codes must take into account the physical and technical limitations of the robot, as well as its available expressive capabilities. These constraints include the size and body shape of the robot, available movement patterns, and facial expressions, among others. On the other hand, nonverbal codes must be able to be perceived by the human sensory system and interpreted in accordance with human experience and expectations. We must avoid nonverbal codes that are not detected by the human sensory system, including meaningless signals and codes that are difficult to interpret due to the person having no prior experience with code, i.e., no existing schema for directing their perception of the stimulus. Therefore, the framework presented here (Figure 1) uses the human sensory channels as a foundation. It shows how a robot could be designed to address each sensory channel with different types of nonverbal codes inspired by human communication studies. The main focus is on vision and auditory sensory channels. However, we also explore how haptics and olfactory sensory channels can be addressed and included the gustatory channel for completeness.

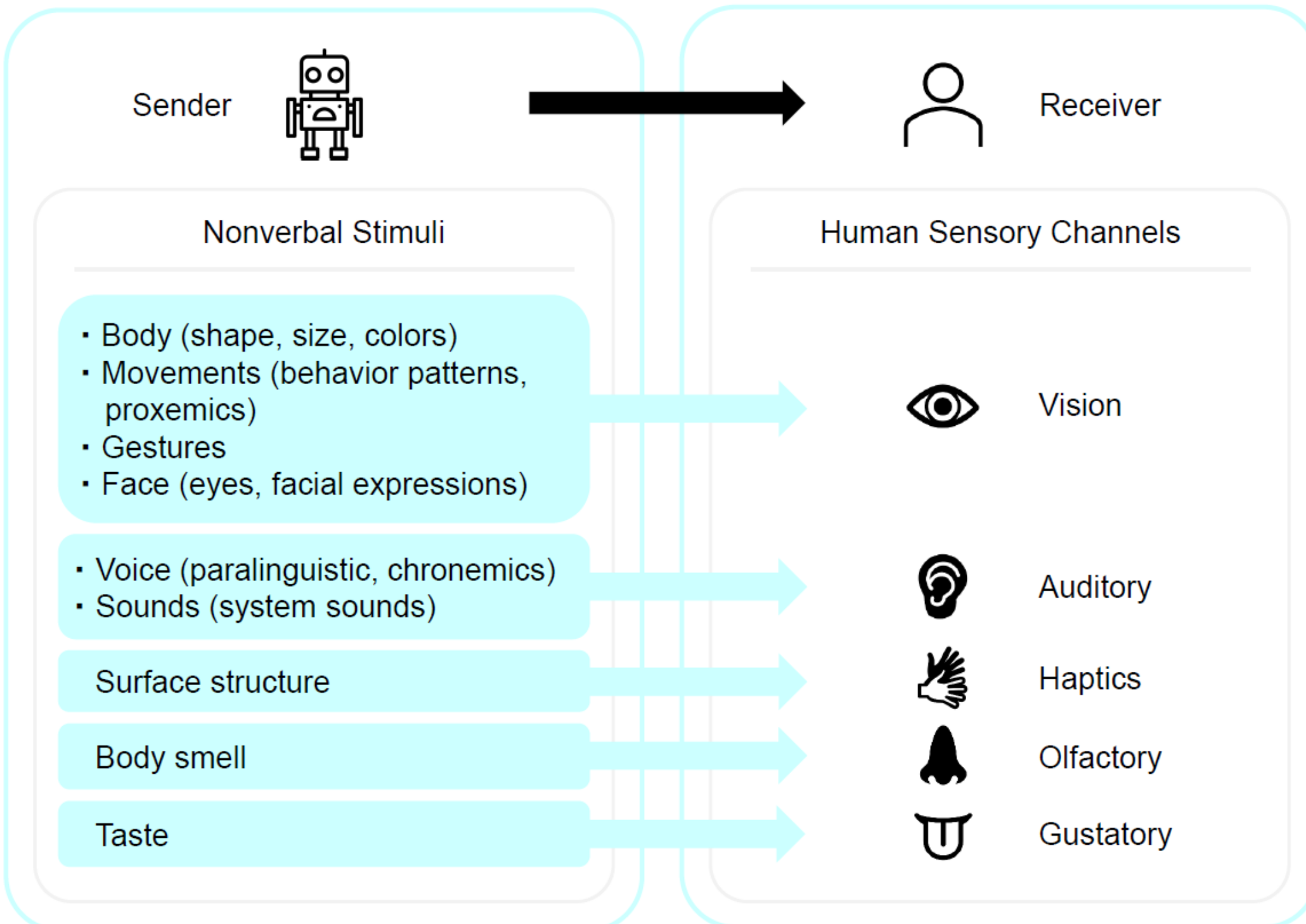

Figure 1: Relation of nonverbal codes for HRI to the five human sensory channels

## 4.1 Visual Nonverbal Codes

### *4.1.1 Physical Appearance and Object Language*

Communication is often preceded by visual observation of the communication partner, including their physical appearance. In HRI, the most relevant examples may be body type, clothing, and object language. Objects or artifacts can transmit powerful messages about the sender. Object language refers to all intentional or unintentional displays of material things [35]. Object properties such as size, shape, form, texture, color are important communication factors. They affect the affective responses of the receiver. For example, symmetrical objects and round shapes with smooth edges stimulate a positive impression, whereas sharp objects can create a formal or negative impression.

#### *4.1.1.1 Body type*

Early research on body type distinguishes between three typical forms: 1) *endomorph*, being short, round, and fat; 2) *ectomorph*, being tall and thin; and 3) *mesomorph*, being muscular and *athletic* [36]. Peoples' aesthetic preferences as well as expectations on a robot's capabilities based on a specific body type must be considered. A wealth of research exists on robot appearance and form factor, including body type; see Fink [37] for a review that remains relevant to this day. As Fink [33] found, robots that are more humanlike in shape tend to perform better in social interactions with people. But this can be a double-edged sword: if humanoid robots are placed in social situations but are not equipped with the behaviours necessary to meet human expectations of the humans that they represent, user experience will suffer. Additionally, the heads and faces of robots will receive the most attention, so nonverbal codes involving these areas of the robot body may deserve the most attention. Another common theme is the relationship between body shape and perceived gender of the robot (e.g., [38, 39]); notably, cultural context can affect interpretations of gender, as found in the cases of Wakamaru [40] and Pepper [41]. What does not exist is a link to communication studies. One exception is the work of Jacquemont and colleagues [42], who applied a communication studies perspective on robotic body types so as to match people's impressions of their temperament. Specifically, they created an explicitly endomorphic robot form, with many round and large components to its shape. However, to the best of our knowledge, results on the actual user experience of this robot have not been explored.

Applying established communication studies frameworks in HRI studies could enhance conceptual rigor and provide a way of doing systematic literature surveys on different body types across different robots, contexts, user groups, and so on. Non-human or proto-human (even alien) forms could be explored, with inspiration drawn from animated characters. For instance, Pokémon, the fictional pocket-holdable monster characters from the series of the same name, are perceived as cute and engaging despite being generally non-humanoid and even non-animaloid in body type.

#### *4.1.1.2 Clothing*

Clothing or dress is another nonverbal code and especially important in creating a first impression. For people, clothing functions as comfort-protection, preserves modesty, but also serves as cultural displays about the social class, gender, age, profession, or status in the hierarchy, to name a few. To our knowledge research has not explored so far how clothing might change people’s perception of a robot. However, we believe cloths could be beneficial especially for robots working in social contexts. Clothing can be quickly created and changed. As such, it may be one of the quickest and easiest nonverbal cues related to physical appearance to explore. Many robots, such as KASPAR [43], have removable clothing, but the use of clothing, varieties of clothing, and even manner of wear, such as topless versus bottomless, has not been well-studied, despite the possibility of important effects according to human models. Future work can start to explore the possibility of designing clothing *for* robots, i.e., robot-centered robot clothing and robot fashion [44]. Furthermore, clothing can be an easy way to clarify the role of a specific robot in public spaces or its affiliation, similarly to how people wear a specific type of uniform or clothing that features the company logo across a variety of jobs.

#### *4.1.1.3 Kinesics: Body Movement and Gestures*

Body movements and gestures are part of the kinesic category of nonverbal cues. These include all nonverbal cues that involve motion of certain parts of the body or the entire body [45], such as emblems, adaptors, body posture, and body language. Various taxonomies for gestures, as a particular type of body movement, have been described in the communication literature [46-48]. For example, a well-used framework by McNeill [46] categorizes gestures as iconic, metaphoric, or deictic. Gesture taxonomies are provisional working tools tailored to particular research interests [46]. Therefore, classification systems for gestures in human communication can only be partially applied to HRI. Additionally, some are limited to certain parts of the body, such as the hands and arms in the case of McNeill [46]. As such, we chose to anchor this work on the taxonomy of Ekman and Friesen [48], which offers the following categories: *emblems*, or gestures that are a direct verbal translation; *illustrators*, or movements associated with speech; *regulators* that regulate speech acts; and *adaptors*, or repetitive movements for self-satisfaction, based on the earlier work of Efron [47]. Efron's categorization system refers to the whole body and thus seems well-suited as a basis for HRI.

#### *4.1.1.4 Emblems*

Emblems are gestures that have a direct literal translation shared by the members of a social group. These are often actions or behaviours that have a certain meaning in certain contexts. Miming gestures, for instance, illustrate a certain action that typically is taken, such as opening a cupboard door or performing a dance move. As such, emblems are often culturally defined, varying over time and across generational cohorts [49]. In most cases, it is better to avoid using such cues in HRI because receivers might have difficulty decoding them. Conversely, robots designed for specific cultural contexts may need to use emblems to be accepted and understood. Nevertheless, while some have recognized that emblems may be culturally sensitive, most have not considered this directly within the context of their own work. We now cover some examples. Cabibihan et al. [50] compared the performance of a person and a robot at performing iconic and emblematic gestures, such as hugging, carrying a baby, and nodding. However, these were not distinguished as iconic, emblematic, or another type of nonverbal code. Carter et. al. [51] built emblematic gestures into the design of a robot that plays catch with a person: throwing up its arms or shrugging on a near miss and shaking its head back-and-forth if the ball was thrown out of range, whereas Gross et al. [52] explored the use of emblematic gestures in place of speech, e.g., using a thumbs up gesture for "great." Zheng, Liu, and Meng [53] explored how people interpret ten "common" emblematic gestures performed by the Nao robot (Aldebaran/SoftBank Robotics), including clapping and handshaking. We may be able to guess what culture(s) are relevant to the results, based on the location of the work conducted. However, this should be stated clearly. Indeed, we should not shy away from exploring emblems in HRI. For example, a robot able to demonstrate mastery of the emblems of a specific sub-culture might be more readily accepted by that group, with its demonstration facilitating its integration into the group or work-team. Therefore, we must be conscious of cultural rules and justify our design decisions with and for the cultural group involved. We recommend taking a human-centered design approach, where the needs, expectations, and desires of the particular cultural and/or generational group are mapped out and then applied to the robot. Participatory and co-design methodologies, as well as iterative testing with members of the cultural group, are examples of how to achieve this.

#### *4.1.1.5 Illustrators*

Illustrators are gestures that provide a description of what is being related vocally, such as the size or shape of an object demonstrated by drawing the outline of the object (iconic gestures), or an indication of the speed of an object by way of fast or slow hand movements. Illustrators are also used to highlight important points in a speech; see Streeck [30] for examples. They may be the most common type of gestures used in human communication. Salem et al. [54] evaluated how well people perceive illustrators, such as the size and shape of a vase, performed by a Honda robot. Adding pointing gestures to a Nao robot delivering vocal directions, improved its performance [55]. To ensure the interpretability of the illustrators used, human gestures can be analyzed in a specific context, and these can then be transferred into a robotic hand, as was done by Sheikholeslami et al. [56]. So far, much work in robotics has focused on developing robots that perceive human gestures (e.g., [57, 58]). More work is needed the other way around: whether people can perceive illustrative gestures performed by robots accurately and the degree to which such gestures are useful. Illustrators add redundancy to speech by encoding what is being said into gestures, thus providing additional cues of how to interpret a message. Work has also started on two-way interaction between people and robots enabled by other technologies, such as wearables (e.g., [59]). This sets the stage for new forms of illustrator-based bidirectional communication.

#### *4.1.1.6 Regulators*

Regulators are body movements and gestures that manage and direct a conversation. Regulators, such as head nods, eye contact, and hand gestures pointing to a specific communication partner (deictic gestures), are used to manage the conversation flow and turn-taking between speakers. Their absence can make conversations in larger groups especially difficult. Consider the case of Zoom meetings where several people try to speak at the same time. Regulators are especially necessary in HRI when more than one person interacts with a robot simultaneously, or when more than one robot interacts with one or more people simultaneously. Some previous research has incorporated regulators into a robot's nonverbal cue set. For example, Sugiyama et al. [60] developed a "facilitation approach" to increasing the effectiveness of deictic pointing gestures with the Robovie humanoid robot. Regulators, such as pointing to direct the actions of a human partner, are especially effective in combination with speech and eye gaze [54], a finding also supported in collaborative tasks [61]. Other examples of regulators used in HRI are hesitation gestures in the form of the robot slightly withdrawing its hand when it and a person reach for the same object at the same time [62], or the robot nodding while maintaining eye contact during back-channeling (verbal interjections) and turn-taking scenarios in response to more or less extraverted people [63]. In general, regulators as gestures are easy to implement into a robot's gesture repertoire. The challenge is that the *timing* of such gestures requires a certain situational awareness. The first stages of implementation could start with a robot using regulators to signal the beginning or end of speech. More advanced systems should be able to manage the flow of the conversation by encouraging a person to respond to specific cues, or even being able to encourage one person out of a larger group of people to join the discussion, making sure everybody contributes evenly. Thus, robots could become mediators in group discussions. They might even be more readily accepted as mediators because of their perceived neutrality to specific topics.

#### *4.1.1.7 Adaptors*

Adaptors (sometimes "adapters") are repetitive gestures and body motions that satisfy physiological and psychological needs. People use adaptors like touching their hair or bouncing their legs to manage stress. Thus, adaptors reveal information about a person's internal emotional state. Adaptors may be conscious or unconscious, controllable or uncontrollable. As such, they could be useful in HRI for improving humanlikeness and a sense of reciprocity, especially in terms of social emotions or emotional contagion. Very few examples on the use of adaptors in HRI exist. Vocal and body adaptors (an "umm" and touching the chin) have been explored in awkward conversation settings [64], or added to increase social appropriateness at the end of a conversation [65]. However, few studies have explored adaptors as part of a robot's regular gesture repertoire. We can imagine many possibilities that should not be technically difficult to implement given what has so far been achieved with other codes. We can work up to a range of small cues working in concert to produce a larger impression intended to reveal a robot's internal state, such as a nervous-looking robot that breathes, shifts side-to-side, clutches its fingers, cowers a little, blinks rapidly, and so on. Robots displaying adaptors could also communicate a certain degree of vulnerability and insecurity that might make them more relatable and approachable, especially in first-contact situations or with inexperienced users. A robot development kit made up of the smallest gestures and behaviours may be useful for crafting a range of adaptors.

#### *4.1.1.8 Posture and Body Language*

Posture refers to a special position of the body or the way that a person holds their body. The four basic body postures are standing, sitting, squatting, and lying down [66]. There are many variations in postures, but a basic distinction can be made between dynamic postures (when the body is in motion, e.g., walking) and static postures (when the body is not moving, e.g., sleeping or standing). Position and posture can communicate the attitude and relationship status of a sender [67]. For example, Mehrabian [68] found that when communicators who were women adopted an open posture, they were interpreted as having a more positive attitude than when they adopted a closed posture. Dynamic postures are also a rich source to express emotions, e.g., shying away when anxious, raising one's arms in an outburst of happiness. Such "emotional body language" involves the whole body and coordinated movements that facilitate the interpretation of a person's emotions [69]. Using the body language of robots to express emotions appears to be an effective tool, as well. Robotic emotional body language can be interpreted similarly to that of humans, and people can correctly identify emotions displayed in key poses [70]. A review of studies on affective-expressive movements for robots by Karg, et.al. [71] revealed that work so far has mainly focused on a small set of movement types restricted to the upper half of the body. They suggest that more research on the relationship between movement type and style of expressiveness is needed, e.g., whether walking up and down, slow or fast, will result in completely different impressions of the robot.

Full body motion, like jumping up and down while waving your hands towards a friend who is trying to find you in a crowd of people, is generally an effective way to attract attention. In HRI, body posture can communicate the activation status of a robot. An upright, straight posture can indicate high activation, while a relaxed, crouched posture can indicate inactivity. Similarly, lifting or lowering the head can indicate activation or inactivation. These behaviours can be seen in the Cozmo and Vector robots (Digital Dream Labs) and the Nao robot. Notably, robotic motion and movement of body parts is often accompanied by the sound of the motors involved in these actions, which can be distracting to the receiver [29]. We can explore whether subtle social human phenomena apply to robots. For instance, a robot can incline towards a speaker or mimic the conversation partner's posture to promote feelings of interest and sympathy [72]. This could have implications for power and social hierarchies, as well as robot acceptance and affect. Therefore, robots taking on different roles could communicate their status via specific postures, such as a robot teacher using power poses [73], e.g., standing upright with legs slightly apart, to create an aura of authority, or a service robot using a submissive pose, e.g., slightly bending its body or leaning towards a customer to communicate respect and attentiveness. Some work has also started to adapt human measures and instruments. In particular, McColl and Nejat [74] applied the Position Accessibility Scale (PAS) from the Nonverbal Interaction States Scale [75] to an HRI context. Others may be adapted for HRI.

#### *4.1.1.9 Proxemics*

Proxemics refers to the perception and management of space, including distance from others. For robots, proxemics is as part of its kinesthetic body language. People create boundaries and control areas of space with their body language. Cultural groups have different rules for what kind of behaviour is accepted depending on the type of territory (e.g., [76, 77]). Certain intimate behaviors that are exhibited in the home might not be acceptable in public, and people are usually expected to follow these rules. This applies to dress codes as well. A special type of perceptual or psychological space is the *personal space*, a kind of invisible bubble that surrounds a person. People are sensitive about violations of personal space, as it defines the acceptable distance to other people based on many contextual factors ranging from gender, relationship status, social hierarchy, and so on. Violations of personal space can trigger strong reactions, such as high arousal, anxiety, and physical aggression [78].

Proxemics has been widely studied within HRI. In an early work, Walters et al. [79] used Eysenck's model of personality traits to explore spatial zones in HRI, finding that people who were more "proactive" kept greater distance compared to others. Stark et al. [80] explored comfort levels when a robot violated the person's personal space during a collaboration activity. People's approach behavior towards robots is influenced by whether they perceive the robot as acting autonomously or as teleoperated [81]. It is also influenced by perceived liking and gaze patterns, such that people keep away from unlikable robots with sharp gazes [82]. Furthermore, previous studies by Kim and Mutlu [83] revealed a relationship between proxemic distance (close versus far), task distance (cooperative versus competitive) and power distance (supervisor versus subordinate), showing that consistency in these various kinds of distancing is key. Peters et al. [84] found that people responded akin to human-human interaction contexts when walking with virtual human characters and a humanoid robot (in this case, Pepper) in terms of social distance.

Many other factors, nonverbal and otherwise, can influence proxemics between people, and this is sure to be true in HRI contexts, as well. The experience of near-future applications on exoskeletons and exorobotics (e.g., [85]), where the robot is intimately close to the human body, will need to be evaluated. While people usually have some understanding about the size of another person’s personal bubble and try to avoid intrusion into this space, a similar awareness might not exist for or by robots. Clear visualizations of a robot's personal space could prevent violent, inappropriate, and/or dangerous inter/actions with and/or towards robots. Another question is whether human-like robots are more likely to be perceived as having something like a personal bubble compared to machine-like robots, or if the size of this perceived space differs. These questions might be especially useful for explaining people's approach and avoidance behaviors towards different types of robots.

#### *4.1.2 Facial Expressions*

Facial expressions can be very complex and are often interpreted as displays of emotions. Facial expressions for basic emotions such as happy, sad, surprised, angry, scared, and disgusted are believed to be universal [86]. However, they are also psychologically-constructed mental events based on internal and external factors to the person [87]. The emotion of "sadness" for one person may not match another, or even match that same person in other events or times. Specific cultural rules apply when it is appropriate to show certain facial expressions. People learn early on how to modify facial displays as part of becoming good communicators. Japanese people, for example, try not to show sadness because it might have a negative effect on others. Similarly, Western cultures often expect women to smile and withhold expressions of anger, while men can go without smiling and their expression of anger tends to be more accepted [88]. Furthermore, we learn rules for facial expressions that are adequate for specific professional environments: the so-called *professional display rules*. Kindergarten teachers learn to appear cheerful and kind, shop assistants smile and are friendly even when confronted with customer complaints, and so on.

A tremendous amount of work has gone into exploring facial expressions in robots. Seminal work includes Breazeal and colleague's Kismet [89], the eye-centered robot by Tojo et al. [90], Kanoh and colleague's Ifbot [91], Hashimoto and team's expressive SAYA [92], and Bennewitz and colleague's Fritz [93], to name a few. More recently, a range of high and low fidelity options have been explored. On the high end, the humanoid headbot Sophia by Hanson Robotics has made waves with its ultra-realistic facial expressions. On the low end, Seaborn et al. [94] used simple LED eye animations combined with nonverbal utterances to convey expressions of pleasure and dismay in a semi-humanoid robotic food waste recycling bin. Current efforts on facial displays in robots have largely focused on making robots more human-like by giving them the appearance of being able to feel emotions. However, the purpose of facial expressions is not limited to revealing one's inner affective state; facial expressions are also an important tool for building relationships with others. Research should explore the purposeful use of facial expressions in different contexts with the goal of creating shared understandings and situational awareness between robots and humans, as well as to communicate feelings of empathy. Robots "pretending" to have emotions are perceived as uncanny by some [95]. Thus, using emotion as a *social* response to certain situations (e.g., a robot that fails to complete a task highlights this failure to a human partner by expressing visible anger) *without* the goal of revealing an inner affective state might be more valuable for HRI settings.

#### *4.1.3 Eye Behavior (Oculesics)*

It is said that eyes are the window to a person's soul. True or not, eye behavior is key to successful social interactions. Eye gaze can promote or hinder contact. Mutual gaze is usually the first step in starting an interaction, signaling awareness of the other person and a willingness to engage with each other [96]. According to Adams and Kleck [97], direct gaze is associated with approach behavior, whereas averted gaze is associated with avoidance behavior. Furthermore, direct gaze can be interpreted either as threatening or friendly, depending on other contextual cues. Recent work suggests that people do not necessarily look into each other's' eyes; instead, eye gaze fluctuates between the eye and mouth region, indicating that *face gaze* may be a more appropriate moniker than "eye gaze" [98]. Eye behavior might also be associated with certain character traits, as suggested by Grumet [99]. For example, the relationship between eye gaze and extraversion was explored in the ECA Linda [100], where extraversion was defined as gazing at the user 90% of the time and introversion was set to 30% of the time; however, the effect of differing eye gaze could not be experimentally detected. Another potential source for social cues is pupillary dilation. The constriction and dilation of the pupil has the primary function of controlling the amount of light admitted into the eye. However, studies suggest that pupil dilation can also be a sign of interest [101]. Furthermore, pupil size can correspond to negative emotional expressions, with smaller pupil sizes indicating higher intensity and sadness of perceived emotions [102, 103].

In HRI, eye gaze is often paired with cues that can be classified as other nonverbal codes, especially regulators; we have already described several (e.g., [54, 61, 63]). Previous research has revealed positive effects of robotic eye contact. For example, a robot who kept eye contact with one member of a group, to the exclusion of others, was seen as favouring that member [104]. Also, shared gaze attention cue during handovers of a bottle of water sped up the handover process [105]. In contrast, van Dijk et al. [106] evaluated eye gaze in combination with verbal messages and gestures, but did not find an effect. Eye gaze has also been explored as a control mechanism. For instance, Yu et al. [107] explored eye gaze for object selection by a human operator during control of a teleoperated drone. Eye gaze has long been of interest in HRI. Future work can explore more complicated situations as well as subtleties in gaze, especially gaze switching at certain moments in an interaction. Furthermore, some topics in eye gaze have received almost no attention in HRI. For instance, pupillary dilation has not been well researched yet. Changes in pupil dilation could provide additional cues to a person interacting with a robot, indicating interest (dilated pupils) or focus and intensiveness (smaller pupils). These are technically easy to implement into a robot. Design inspiration can be drawn from human models as well as from fictional representations of human(oid) characters in comics and animation.

## 4.2 Auditory Nonverbal Codes

### *4.2.1 Paralanguage*

Auditory codes are another human sensory channel sensitive to nonverbal codes. *Paralinguistics* is a key element of this phenomenon. Paralinguistics, also called vocalics, is the study of *paralanguage*, which refers to the sounds that accompany speech but are not words. This includes *vocal qualities*, such as pitch, tone and intonation patterns of the voice; *vocalizations*, from groans to giggles to cries to yawns; and *vocal segregates*, such as vocal fillers like "uh", "ah", "um," pauses, and dragging out sounds at the start or end of words. Paralanguage can convey affect, such as how laughing out loud conveys being happy and yawning conveys feeling tired or bored. Silence is also an element of paralanguage. Silence can have a positive or negative meaning. It can convey respect, intimacy, or comfort. On the other hand, silence when a response is expected can generate a negative impression or signal rejection, such as not laughing when a joke is told or not replying to a question. In general, people are very sensitive to the emotional quality of a voice and can detect if a person is happy or sad, cynical or sarcastic. Voice parameters indicating different emotions include pitch, pitch range, and pitch change, voice intensity, tone and contour [108].

Paralanguage has a long and rich history in robotics, stemming from early explorations in computer voice [109]. See Seaborn et al. [5] for a review featuring many examples across the robotic spectrum. Much work has focused on replicating speech, i.e., robots made to speak human languages. Recently, there has been a turn to non-speech, or at least speech that does not involve words: *non-linguistic utterances* [110]. In an early paper, Read and Belpaeme [111] described non-linguistic emotive auditory cues for child-robot interaction. Savery et al. [112]

created Shimi, a robot that sings without words and moves without humanoid limbs to convey emotionality. Non-linguistic utterances and other forms of vocal but non-speech paralanguage can get around the sticky issue of natural language processing (NLP). Another promising direction is robotic regulators for *backchannelling*, or verbal interjections, and its various forms [113]. Backchannelling can be very frequent, as in aizuchi (interjections to indicate that the listener is paying attention) during Japanese conversations, to less frequent, as well as being comprised of a mix of verbal and nonverbal cues. Certain cues, such as laughter, are virtually uncharted territory in the design of robots, even though there are robots that detect and respond to laughter (e.g., [114]), as well as much work on using human laughter to measure engagement and enjoyment in HRI contexts (e.g., [115]).

#### *4.2.2 Chronemics*

Chronemics is the study of the use of time in nonverbal communication. People can have different time orientations, being focused more or less on the past, present, or future [116]. Time orientation can influence the cognitive level of behavioral functioning. Beliefs grounded in past experiences, current appraisals, and reflections on future options affect people's actions and motivation [117]. A person's time orientation can affect how they perceive time as well as their willingness to wait or be punctual. Chronemics, as such, can be expressed through body language and behavioral patterns. Since chronemics has been studied mainly in conversational settings, we have included it in the category of auditory nonverbal codes. Time distribution in conversation is an expression of each person’s power, shown through the length of the conversation, turn-taking, speech time, and conversation initiation and closure.

Most of the research on chronemics has focused on conversation scenarios, notably turn-taking (e.g., [118, 119]), conversational fillers (e.g., [64]), and topical time orientation [120]. Time orientation poses several relevant questions for robots. Do robots have a time orientation? Or are they perceived to have a time orientation by people? Is a robot speaking quickly perceived differently from a robot speaking slowly? Should a robot pause when it is processing information? Should a robot be late in a culture where people are usually not on time? Could robots amplify a person's time orientation by preserving the past and extending one's existence into the future? For example, the Japanese artist Etsuko Ichihara[3] created a robot with the printed face of a deceased person, imitating the person's speech and behavior. Inspired by traditional Japanese funeral rites, the robot spent 49 days (called shijukunichi, the time period during which a spirit is expected to reach the next life, according to Shinto traditions) with family members to gradually bid farewell. Taxonomies of specific varieties of chronemics may be created or adapted for developing HRI scenarios. For instance, the taxonomy of conversational interruptions developed for ECAs by Cafaro, Ravenet, and Pelachaud [121] could be translated to HRI contexts. Indeed, their procedure of involving people choosing the optimal behaviours, i.e., nonverbal cues in response to interruptions, to develop the underlying algorithm could be replicated with robots.

[3] Learn more about Etsuko Ichihara's "Digital Shaman Project" here: https://etsuko-ichihara.com/works/digital_shaman_project/

### 4.3 Haptic Nonverbal Codes

Haptics refers to somatosensorial sensation. Haptic communication is how people and other animals communicate via touch. As such, haptics might be the most primitive but also the most essential form of nonverbal communication. For example, studies have shown that foster children deprived of touch have developmental delays [122]. The *affective touch hypothesis* proposes that touch has an important effect on emotional well-being [123]. Positive emotions evoked through gentle strokes (e.g., grooming) seem to mediate socializing behavior [124] and appear to play a crucial role in autism [125]. Furthermore, touch deprivation is linked to depression and violence [126, 127]. On the other hand, cultural groups have different rules about who, how, and how often members can touch each other. Violations of these rules can have social and legal consequences, such as touch or body contact that is perceived as being inappropriate or unwanted. Touch initiation is a dominant behavior that controls or directs the behavior of others. For instance, people perceived to be of higher status may initiate touch. However, touching also fulfills basic psychological needs. Hugging friends and family members or stroking the warm fur of animals creates positive feelings, is comforting, and reduces stress.

Haptics has long been explored in HRI, like the seal robot by Shibata and Tanie [128] that can be held and stroked, or the pet alternative robot by Yohanan and MacLean [129]. Other research has looked into surface temperature, finding that warm temperature was important when people held hands with a robot [130]. Hu and Hoffman [131] created a robot that can have goosebumps. Robots that respond to touch might provide an alternative approach to therapy for people with autism, a developmental disorder associated with various sensory-perceptional anomalies [132], as suggested in a recent systematic review on social robots interacting with autistic people [133]. On the other hand, touch in human interaction fulfills different needs but also follows social rules. We must then recognize the ethical implications of increasingly realistic touch sensations and situations in HRI [134]. Would people accept being poked by a robot to grab their attention, for instance? There are many questions to explore. Surely robots should respect existing cultural rules about touching. A robot initiating touch might be perceived as intruding into someone's personal space, or even an aggressor, depending on the context and expectations. On the other hand, a robot providing haptic feedback when touched by a person could foster intimacy and support relationship-building, an important aspect for social robots.

### 4.4 Olfactory Nonverbal Codes

Olfactics, or the sense of smell, is probably the least understood of all human sensations. People seem to have universal preferences for scents that may have biological and evolutionary roots. Qualities and thoughts attributed to different scents seem to play an important role. For instance, we may feel that people who smell "bad" are repellent, while people who smell "good" are attractive. Olfactory sensation seems to have an influence on social interaction as well. For example, humans seem to generate specific chemo-sensory signs to signal a specific emotional state, such as fear and happiness, and some people can fairly accurately distinguish these [135]. De Groot et. al. [136] compared mood odors for disgust and fear, showing that they produced matching facial expressions in participants who were exposed to both of these odours. There is also growing evidence that every person has a unique odour signature [137] that can be identified by close relatives [138]. However, olfactics has been all but neglected in HRI so far. Even so, we believe that this is an area ripe for research. Some work in HRI [139] has explored robots that are able to detect and distinguish scents, especially in teleoperation contexts such that the robot's "nose" is an extension of the human operator's senses (e.g., [140]). Work in the related field of HCI has explored olfactory displays (e.g., [141, 142]), providing technical proofs-of-concept. One project combined robots and an olfactory display in a 4D theatre experience [143]. Yet, HRI work on scent is virtually nonexistent. We can translate this work to HRI. We can explore what scents are suitable for which kinds of robot forms. We can examine whether and how people may be influenced by different scents in subtle ways. For instance, can a robot's scent affect approach distance or delineate its personal bubble? We can evaluate the relationships between scent and other factors, such as personality and humanlikeness. Specific scents could also help to personalize robots and make them unique. There is much work to be done.

### 4.5 Gustatory Nonverbal Codes

The gustatory sense, or sense of taste, is not a nonverbal code usually considered in the communications literature. However, we include it to complete the list of the human sensory channels involved in HRI. Taste receptors can be mainly found on the human tongue. Research has shown that we can distinguish five different taste modalities: sweetness, sourness, saltiness, bitterness, and savouriness (also called umami) [144, 145]. In most HRI contexts, the gustatory sense will certainly not be relevant. However, robots used for feeding people, such as robots that assist young parents, robots in care facilities for older people, or robots for patient care, could stimulate patients' gustatory senses during feeding. In general, gustatory sensation weakens in older people [146], giving the impression that food is tasteless and consequently reducing appetite. Also, certain medical conditions, such as "long COVID," involve a loss of taste [147]. Some research has experimented with taste displays that can reproduce the five basic taste modalities using gels [148, 149]. Integrating this technology into feeding robots could enhance or alter the taste of food according to patient preferences to stimulate appetite and ensure patients receive the necessary amount and balance of nutritious food.

## 5 CONCLUSION

We have provided a comprehensive overview of nonverbal codes identified in human-human communication by the field of communication studies and outlined how these nonverbal codes appeal to the human sensory systems involved in HRI, including vision, auditory perception, haptics, olfactory, and gustatory perception. We have translated these to the case of robots, offering several suggestions for research and development in robotics and HRI. We believe that the nonverbal codes we have covered in this paper are especially promising for HRI research. While some topics have been explored already, notably body movement and gestures, other topics, such as haptics, olfactics, and chronemics, are ripe for exploration. Methodologically, this was a difficult paper to compile because much of the HRI literature does not use the terminology of communication studies. For instance, we found references to "deictic" as a sub-category of the nonverbal code "regulators," but the term "regulators" is not mentioned. Additionally, while there is much work to be found in HRI under certain cross-disciplinary terms, such as "gestures," these terms are far too broad to allow easy navigation and categorization of that literature. As such, we could only provide a sampling of examples based on the keywords used in HRI papers and the literature that we know as researchers working within or adjacent to HRI. We also aimed to include work from a variety of authors covering different regions, fields of study, and career stages so as to capture a range of ideas from the diversity of people working in this area. As such, the work presented here offers a mapping of how a communication studies framework can and does apply to HRI. Let us begin the process of interdisciplinary categorization and chart a path for future research together.

## ACKNOWLEDGMENTS

Our sincere gratitude to the reviewers. This work was supported in part by the Japan Society for the Promotion of Science (JSPS) through a KAKENHI Wakate grant (#21K18005).